\documentclass[11pt,a4paper]{article}
\usepackage{fontspec}
\usepackage[a4paper,textwidth=178mm,top=18mm,bottom=19mm]{geometry}
\usepackage{amsmath,graphicx,booktabs,tabularx,array,microtype,xurl,caption,stfloats,xcolor}
\DeclareFontShape{OML}{cmm}{b}{it}{<5><6><7><8>gen*cmmib<8.5->cmmib10}{}
\DeclareSymbolFont{rapdboldletters}{OML}{cmm}{b}{it}
\DeclareMathSymbol{\rapdtheta}{\mathord}{rapdboldletters}{"12}
\DeclareMathSymbol{\rapdphi}{\mathord}{rapdboldletters}{"1E}
\usepackage{flushend}
\usepackage[numbers,square,sort&compress]{natbib}
\usepackage[hidelinks,unicode]{hyperref}
\DeclareMathSizes{10.95}{12}{8}{6}
\DeclareMathSizes{7.5}{8}{6}{5}
\AtBeginDocument{}
\newcolumntype{Y}{>{\raggedright\arraybackslash}X}
\DeclareCaptionLabelFormat{continuedtable}{#1~#2 (Continued)}
\newcommand*{\declarationheading}[1]{\section*{#1}}
\newcommand*{\declareifset}[2]{\ifx#2\empty\else\declarationheading{#1}#2\par\fi}

\newcommand*{\CorrespondenceEmail}{tainchi.lu@polyu.edu.hk}
\newcommand*{\AuthorContributionsText}{Shangkun Sima designed the methods, curated the data, developed the software and computational workflows, performed the analyses and validation, created the figures, and drafted and revised the manuscript. Tongfei Shen participated in primary-source data provenance verification and reconciliation and edited selected passages of the manuscript, refining their wording to improve clarity and readability. Tianchi Lu formulated the initial research question, supervised the study, and made substantive revisions to the final manuscript.}
\newcommand*{\AcknowledgementsText}{We thank the ProNAB team at the Indian Institute of Technology Madras for curating binding-affinity data, the Worldwide Protein Data Bank consortium and structure depositors for making structural data available, and the researchers and institutions that generated the experimental measurements used in the main and external evaluations. We thank the developers of the evaluated models and supporting software for making their resources available, and the M2-PRNet authors for providing three-view images and preprocessing materials.}
\newcommand*{\FundingText}{This research received no specific funding.}
\newcommand*{\ConflictOfInterestText}{The authors declare no competing interests.}
\newcommand*{\AIAssistanceText}{The authors used a large language model (Claude Opus 4.8, Anthropic) solely to assist in writing exploratory data-preparation scripts and their unit tests in May 2026, a use permitted under the Policy for Acceptable Use of Large Language Models of \textit{Briefings in Bioinformatics}. The assisted code produced no results reported in this paper. Claude Opus 4.8 was not used to draft, rewrite, translate or polish the manuscript. The authors reviewed and tested the assisted code and take full responsibility for it and for the manuscript. Details are given in Supplementary S13.}

\newcommand{\ManuscriptAuthors}{%
  {\fontsize{10.5}{12.5}\selectfont\bfseries
    Shangkun Sima\textsuperscript{1,\textdagger},
    Tongfei Shen\textsuperscript{2,3,\textdagger},
    Tianchi Lu\textsuperscript{4,*}\par}
  \vspace{4pt}
  {\fontsize{8.5}{10.3}\selectfont
    \textsuperscript{1}School of Mathematics and Statistics, Lanzhou University, Lanzhou 730000, China\par
    \textsuperscript{2}School of Information Science and Engineering, Qingdao Huanghai University, Qingdao, China\par
    \textsuperscript{3}Department of Computer Science, City University of Hong Kong, Kowloon, China\par
    \textsuperscript{4}The Hong Kong Polytechnic University, Hung Hom, Kowloon, Hong Kong\par
    \textsuperscript{*}Corresponding author: Tianchi Lu, The Hong Kong Polytechnic University, Hung Hom, Kowloon, Hong Kong. E-mail: \href{mailto:\CorrespondenceEmail}{\CorrespondenceEmail}\par
    \textsuperscript{\textdagger}Shangkun Sima and Tongfei Shen contributed equally to this work.\par}}

\hypersetup{
  pdftitle={Uneven gains in protein--RNA affinity and mutation-effect prediction: mechanisms and reference-anchored local recovery},
  pdfauthor={Shangkun Sima, Tongfei Shen, Tianchi Lu}
}

\newcommand{\figurealt}[1]{%
  \par\noindent{\scriptsize\color{black!75}\raggedright\textit{Alt text:} #1\par}}
\DeclareCaptionFont{preprintnote}{\rmfamily\fontsize{8.5}{10.3}\selectfont}
\makeatletter
\renewcommand{\normalsize}{%
  \rmfamily\@setfontsize\normalsize{10bp}{12.5pt}%
  \abovedisplayskip 8pt plus 2pt minus 3pt
  \belowdisplayskip \abovedisplayskip
  \abovedisplayshortskip 3pt plus 2pt
  \belowdisplayshortskip 5pt plus 2pt
  \let\@listi\@listI
}
\renewcommand{\section}{\@startsection{section}{1}{\z@}%
  {-2.1ex \@plus -.5ex \@minus -.2ex}{.8ex \@plus .2ex}%
  {\normalfont\large\sffamily\bfseries}}
\renewcommand{\subsection}{\@startsection{subsection}{2}{\z@}%
  {-1.55ex \@plus -.4ex \@minus -.2ex}{.55ex \@plus .15ex}%
  {\normalfont\normalsize\sffamily\bfseries}}
\makeatother
\DeclareMathSizes{10bp}{9}{7}{5}
\DeclareMathSizes{9}{9}{7}{5}
\DeclareMathSizes{8.5}{8}{6}{5}
\DeclareMathSizes{8}{8}{6}{5}

\newcommand{\evRapdAbs}{1.1431}
\newcommand{\evRapdOrd}{0.7704}
\newcommand{\evRapdLocal}{0.3660}
\newcommand{\evRapdRank}{0.3123}
\newcommand{\evRapdSD}{0.4700}

\newcommand{\evZeroLocal}{0.3787}

\newcommand{\evZeroSD}{0.0000}

\newcommand{\evControlLocal}{0.3991}
\newcommand{\evControlRank}{0.1886}

\newcommand{\evOfficialAbs}{1.6776}
\newcommand{\evOfficialOrd}{0.6304}
\newcommand{\evOfficialLocal}{0.4244}
\newcommand{\evOfficialRank}{-0.1065}
\newcommand{\evOfficialSD}{0.0016}

\newcommand{\evUoneLocal}{0.3727}
\newcommand{\evUoneRank}{0.2052}

\newcommand{\evAAOrd}{0.8192}
\newcommand{\evAALocal}{0.3749}
\newcommand{\evAARank}{0.1919}

\newcommand{\evMacroLocal}{0.3730}

\newcommand{\evCloudNplSingleLocal}{0.3811}
\newcommand{\evCloudNplSingleRank}{0.5048}

\newcommand{\evCloudNplMeanLocal}{0.3775}
\newcommand{\evCloudNplMeanRank}{0.5810}

\newcommand{\evRapdNplSingleLocal}{0.4176}

\newcommand{\evRapdNplMeanLocal}{0.4075}

\newcommand{\evCloudSfLocal}{0.4144}

\newcommand{\evOldSfLocal}{0.5303}

\newcommand{\evRapdExtAbs}{0.5282}
\newcommand{\evRapdExtOrd}{0.7207}
\newcommand{\evRapdExtLocal}{0.4151}
\newcommand{\evRapdExtRank}{0.0950}

\newcommand{\evOfficialExtAbs}{1.4606}

\newcommand{\evOfficialExtLocal}{0.4241}
\newcommand{\evOfficialExtRank}{-0.2112}
\newcommand{\evOfficialExtSD}{0.0149}
\newcommand{\evControlExtAbs}{0.5282}
\newcommand{\evControlExtOrd}{0.6995}
\newcommand{\evControlExtLocal}{0.3761}
\newcommand{\evControlExtRank}{0.1454}

\newcommand{\evZeroExtLocal}{0.4230}

\newcommand{\evAdditiveExtAbs}{0.6932}
\newcommand{\evAdditiveExtOrd}{0.8845}
\newcommand{\evAdditiveExtLocal}{0.4198}

\newcommand{\evRapdDiffOrd}{-0.0213 [-0.0418, +0.0005]}
\newcommand{\evRapdDiffLocal}{-0.0331 [-0.0569, -0.0120]}
\newcommand{\evRapdDiffRank}{+0.1237 [+0.0109, +0.2596]}

\newcommand{\evUoneOfficialExtSD}{0.0132}

\newcommand{\evPrimaryPairs}{338}
\newcommand{\evPrimaryLocalPairs}{220}
\newcommand{\evOfficialEndpoints}{95}
\newcommand{\evOfficialPairs}{53}
\newcommand{\evOfficialLocalPairs}{35}
\newcommand{\evExternalEndpoints}{27}
\newcommand{\evExternalPairs}{24}
\newcommand{\evLocalFamilies}{20}

\newcommand{\evFrozenOrd}{0.6840}
\newcommand{\evFrozenLocal}{0.3991}

\newcommand{\evFrozenSD}{0.6428}
\newcommand{\evSyncedOrd}{0.6726}
\newcommand{\evSyncedLocal}{0.4004}

\newcommand{\evSyncedSD}{0.7033}
\newcommand{\evSyncedRankDiff}{+0.0023 [-0.0444, +0.0498]}
\newcommand{\evOfficialReductionAbs}{$+0.4237$}
\newcommand{\evOfficialReductionOrd}{$-0.0115$}
\newcommand{\evOfficialReductionLocal}{$+0.0001$}
\newcommand{\evMatchedControlAbs}{1.5154}

\newcommand{\evMatchedControlOrd}{0.5741}

\newcommand{\evMatchedControlLocal}{0.4176}

\newcommand{\evMatchedControlRank}{0.1553}

\newcommand{\evMatchedControlSD}{0.3487}

\newcommand{\evGeometryDiffOrd}{-0.0121 [-0.0185, -0.0057]}
\newcommand{\evGeometryOrd}{0.8071}

\newcommand{\evGeometryLocal}{0.3731}
\newcommand{\evGeometryDiffRank}{+0.0770 [-0.0097, +0.1965]}
\newcommand{\evGeometryRank}{0.2689}
\newcommand{\evAdditiveLocal}{0.3749}
\newcommand{\evUoneOfficialExtRank}{-0.2103}
\newcommand{\evControlReductionSep}{+0.0891}
\newcommand{\evControlReductionNpl}{+0.0815}
\newcommand{\evControlReductionSf}{-0.0396}
\newcommand{\evControlSfRank}{-0.2619}
\newcommand{\evAdditiveSfRank}{+0.7024}
\newcommand{\evIdentityBound}{$2\times10^{-15}$}
\newcommand{\evInterfaceMedian}{67}

\newcommand{\evSyncedSDInterval}{[+0.0164, +0.1287]}

\newcommand{\evTrainLocalRange}{0.1133--0.1443}
\newcommand{\evHoldLocalRange}{0.3568--0.3576}
\newcommand{\evTrainRankRange}{0.813--0.883}
\newcommand{\evHoldRankRange}{0.077--0.131}

\newcommand*{\ReleaseArchiveText}{The baseline audit code, figure sources and cohort manifests are archived at \url{https://github.com/simshk2023/protein-rna-affinity-audit/releases/tag/v2026.11.17}. The authors' original code is licensed under the MIT License. The RAPD assembly and verification scripts accompany this manuscript.}

\begin{document}
\twocolumn[{%
\begin{minipage}{\textwidth}
\sffamily\raggedright
{\fontsize{9}{11}\selectfont\bfseries Problem Solving Protocol\par}
\vspace{5pt}
{\fontsize{16.5}{19}\selectfont\bfseries
Uneven gains in protein--RNA affinity and mutation-effect prediction:
mechanisms and reference-anchored local recovery\par}
\vspace{6pt}
\ManuscriptAuthors
\vspace{7pt}
{\fontsize{10}{12}\selectfont\bfseries Abstract\par}
\vspace{2pt}
{\fontsize{8.8}{10.7}\selectfont\noindent
Across eleven protein--RNA prediction method families, gains in aggregate affinity transferred unevenly to substitutions within a binding system. For controlled CoPRA (retrained from fresh task modules without PRI30k complex pretraining), two error decompositions traced gains to shared endpoint placement and landscape means, with no corresponding reduction in centered local error. An independent audit of Official Released CoPRA found similarly compressed local responses, and PRA-Pred feature tracing and fixed-weight M2-PRNet controls located attenuation within the prediction pipelines. Difference-weighted supervision fitted training-family variation but transferred poorly. Selecting by centered mean absolute error (cMAE) favored zero response, and summing a single-endpoint potential over the interface diluted site-level effects. Reference-Anchored Physical Decoupling (RAPD) retains the controlled absolute-affinity output (mean absolute error [MAE] \evRapdAbs{} pKd) and learns a local perturbation from denoised structural contrasts anchored to the wild type (WT). Across \evPrimaryLocalPairs{} pairs in \evLocalFamilies{} unseen families, RAPD reduced local cMAE from \evControlLocal{} to \evRapdLocal{} pKd, below the additive (\evAdditiveLocal{}) and zero-change (\evZeroLocal{}) references, and raised position Spearman to \evRapdRank{} (paired change \evRapdDiffRank{}). Across \evExternalPairs{} pairs in three independent source-matched families, transfer varied with the structural input and binding interface.\par}
\vspace{4pt}
{\fontsize{7.4}{9}\selectfont\noindent\textbf{Keywords:} protein--RNA binding affinity; mutation effect prediction; benchmark audit; error decomposition; structural bioinformatics; physical decoupling\par}
\vspace{5pt}
\noindent\rule{\textwidth}{0.6pt}\par
\vspace{8pt}
\end{minipage}}]
\section{Introduction}
\enlargethispage{-4\baselineskip}
Single amino-acid substitutions can alter protein–RNA recognition, regulation and molecular function~\citep{amrane2014,dominguez2010}. Their effects on binding affinity follow an exact relationship: \(\Delta\Delta G=\Delta G_{\mathrm{mut}}-\Delta G_{\mathrm{WT}}\), or \(\Delta\mathrm{pKd}=\mathrm{pKd}_{\mathrm{mut}}-\mathrm{pKd}_{\mathrm{WT}}\). Recognition combines base-specific contacts, stacking, electrostatics and conformational changes~\citep{corley2020}. Each binding free energy reflects the balance between the complex and its unbound partners. Accurate predictions at both matched endpoints therefore recover the mutation effect. Whether improvements in aggregate affinity prediction deliver that accuracy is an empirical question. Equilibrium and concentration controls affect both absolute affinities and relative specificity, making matched experimental context essential for interpreting mutation effects~\citep{jarmoskaite2020}.

Kappel et al.~\citep{kappel2019} directly tested this connection: GLM-Score's absolute-affinity advantage did not transfer to relative prediction, motivating Rosetta-Vienna RNP-\ensuremath{\Delta}\ensuremath{\Delta}G and explicit bound/unbound energetics. CoPRA later evaluated protein substitutions with complex-pretrained networks and mutation-effect fine-tuning, including per-complex metrics~\citep{han2025}. These studies establish the problem and motivate examining where gains arise across contemporary method configurations.

\begin{figure*}[!t]
\centering\includegraphics[width=\textwidth]{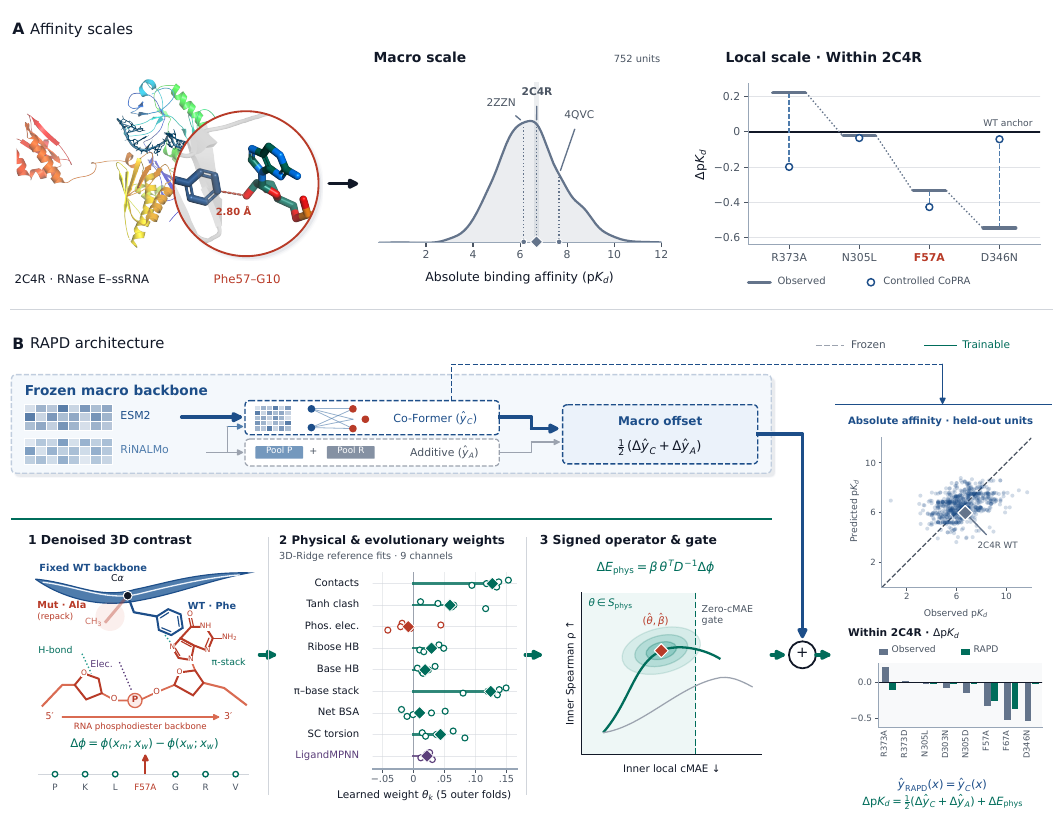}
\caption{\textbf{Binding affinity across systems and mutation responses within a system.}
\textbf{A}, The 2C4R RNase E--ssRNA inset highlights the 2.80~\AA{} distance from Phe57 CZ to G10 O2$^\prime$. The center shows a kernel density estimate of observed absolute affinity for 752 held-out units, with three matched wild-type endpoints marked. The local plot shows four WT-matched 2C4R substitutions ordered by observed mutant-minus-WT $\Delta\mathrm{p}K_d$. Slate segments are observations and open blue circles are Controlled CoPRA endpoint contrasts. The brick tick label identifies F57A.
\textbf{B}, RAPD combines a frozen macro offset with a WT-anchored physical correction. The schematic at left illustrates the fixed peptide backbone, a repacked mutant side chain, and the RNA backbone. Open points show the five outer-fold Decoupled 3D-Ridge reference weights and diamonds show their means. The signed-operator gate is schematic. The upper readout plots observed and predicted $\mathrm{p}K_d$ for the same 752 held-out units against the identity line. The slate diamond marks the 2C4R WT. The lower readout pairs observed (slate) and RAPD-predicted (green) $\Delta\mathrm{p}K_d$ for all eight matched 2C4R substitutions in descending observed order. Each prediction is averaged over three training seeds.}
\label{fig:framing}
\figurealt{The 2C4R ribbon and Phe57--G10 contact introduce the observed affinity distribution across systems and measured responses for four 2C4R substitutions. The RAPD diagram shows the frozen macro backbone, three steps for physical correction, five-fold reference weights, a held-out parity plot, and paired bars for eight 2C4R substitutions.}
\end{figure*}

The distinction extends beyond absolute versus relative error. Shared endpoint errors cancel in a WT–mutant difference, whereas unequal errors remain. Within a mutation landscape sharing one WT, a predictor can improve the average effect without resolving differences among substitutions. Ranking and response amplitude reveal whether such behavior reflects weak responses, incorrect ordering or inaccurate effect sizes (Figure~\ref{fig:framing}A, B).

Using source-linked ProNAB measurements~\citep{harini2022}, we compare eleven method families spanning sequence, structure, multimodal and direct-effect prediction (Table~\ref{tab:model-coverage}), alongside a separate overlap-screened evaluation of Official Released CoPRA. Common-member comparisons and documented experimental constructs connect absolute affinity, ordinary mutation effects and centered variation within landscapes. Two exact error decompositions identify where predictive gains accrue, while PRA-Pred feature tracing and fixed-weight M2-PRNet controls locate attenuation within the calculation of mutation responses. We then test three interventions that change how local effects are learned, selected and represented: a matched three-arm continuation, cMAE-selected contact and hybrid readouts, and a single-endpoint interface potential. These tests motivate Reference-Anchored Physical Decoupling (RAPD), which learns local responses from wild-type-anchored structural contrasts while retaining the absolute-affinity interface. We show that contemporary affinity gains concentrate in shared endpoint placement and landscape means, and that decoupling affinity scale from local physical perturbations recovers substitution ordering in unseen families. Evaluation in three independent source-matched families identifies where that recovery transfers across structural contexts.

\newpage
\section{System and methods}
\subsection{Data and evaluation units}
\begin{table*}[!b]
\centering
\caption{\textbf{Model configurations and matched coverage.}}
\label{tab:model-coverage}
\begingroup
\fontsize{8.5}{10.5}\selectfont
\setlength{\tabcolsep}{4.5pt}
\renewcommand{\arraystretch}{1.05}
\begin{tabularx}{\textwidth}{@{}p{34mm}Ylrrr@{}}
\toprule
\textbf{Method} & \textbf{Configuration} & \textbf{Source} & \textbf{Abs.} & \textbf{Ord.} & \textbf{Local} \\
\midrule
Controlled CoPRA & Controlled family-fold OOF & \citep{han2025} & 752 & 338 & 220 \\
Official Released CoPRA & Released ensemble, no detected PRA310 overlap & \citep{han2025} & \evOfficialEndpoints{} & \evOfficialPairs{} & \evOfficialLocalPairs{} \\
DeePNAP & Controlled family-fold OOF & \citep{pandey2024} & 562 & 242 & 161 \\
OmniBioTE & Official checkpoint and adapted frozen-head OOF & \citep{chen2026} & 736 & 328 & 220 \\
PRdeltaGPred & Structural port (WT and mutants) & \citep{hong2023} & 286 & 235 & 153 \\
GBNL & Author-approved reconstruction (OOF) & \citep{zia2026} & 391 & 149 & 82 \\
SVSBI & Author-approved reconstruction (OOF) & \citep{shen2023} & 391 & 149 & 82 \\
ProRB & Author-provided checkpoint & \citep{xue2026} & 391 & 149 & 82 \\
PANTHER & Native service score & \citep{aletayeb2026} & --- & 87 & 31\textsuperscript{a} \\
PRA-Pred & ss+str and ds+enz applications & \citep{harini2024} & 121 & 106 & 78 \\
mCSM-NA & Direct mutation-effect service & \citep{pires2017} & --- & 246 & 157 \\
M2-PRNet & Five released checkpoints with reconstructed, index-aligned inputs & \citep{wang2026} & 257 & 211 & 131 \\
\bottomrule
\end{tabularx}
\par\vspace{3pt}
\noindent\begin{minipage}{\textwidth}\fontsize{8}{10}\selectfont
Coverage counts absolute units, ordinary pairs and local pairs. OOF means out-of-fold. \textsuperscript{a}PANTHER rank covers 31 pairs in four common landscapes. See Supplementary S2 for implementation details and S5 for matched comparisons.\par
\end{minipage}\endgroup
\end{table*}
We grouped ProNAB records by construct, study and conditions, expressing dissociation constants as pKd = $-\log_{10}(K_d/\mathrm M)$ and taking the median across repeated measurements of one unit. A pair linked one protein substitution to a WT under the same study and conditions. Multiple substitutions, simultaneous protein/RNA changes and unresolved WT links were excluded. Local evaluation required at least five distinct mutated positions after retaining common prediction coverage.

Source reconciliation excluded seven censored units and collapsed ten duplicate Hfq pairs to the source-consistent WT anchors. The principal scopes comprise 752 absolute units, 338 ordinary pairs and 220 local pairs at 203 positions in 21 landscapes and 20 families. Supplementary S1 documents exclusions and experimental-to-input correspondence. Training retains the original fitted membership. Source systems include Hfq, Trm5, Unkempt, Dis3l2 and Dicer~\citep{wang2015,christian2010,murn2016,faehnle2014,tian2014}.

The absolute cohort included reported dissociation constants with usable units and identified constructs. RNA variants could contribute to the absolute cohort. For mutation effects, a landscape groups substitutions sharing an experimental WT anchor. Landscapes with fewer than five positions still contribute to the ordinary mutation summary. In linked absolute analyses, each distinct WT unit is counted once even if it anchors several pairs. Source reconciliation also checks the correspondence between the assayed construct and the mapped protein and RNA chains (Supplementary S1 and Table S30).

\subsection{Models, training data and evaluation design}
The principal CoPRA configuration~\citep{han2025} used frozen ESM2 and RiNALMo encoders~\citep{lin2023,penic2025}, freshly initialized Co-Former and affinity-readout modules, and absolute-affinity mean squared error (MSE) training. Its additive comparator pooled protein and RNA representations separately and summed two readouts with a learned bias. Both recipes shared data, encoder states, structural input selection and optimization settings; their task modules and capacities differed. Protein families, defined by single-linkage sequence relationships meeting 40\% identity and 70\% coverage, remained within one of five outer folds. Each method was fitted at three seeds. Each pair's endpoints used the same fold-specific checkpoint.

CoPRA retained WT coordinates and placed each substitution on the first compatible matched chain. Its official parser used the first coordinate model and a 256-node interface patch. M2-PRNet used five released checkpoints with complete residue-index alignment. Table~\ref{tab:model-coverage} lists all configurations. Supplementary S2 documents mutant-building, preprocessing and training following DOME reporting principles~\citep{walsh2021}.

Official Released CoPRA~\citep{han2025} averages the endpoint predictions of five published checkpoints pretrained on PRI30k complexes and fine-tuned on PRA310. We converted their free-energy outputs to pKd at 298.15\,K and evaluated the ensemble by fixed-weight forward passes on \evOfficialEndpoints{} absolute units, \evOfficialPairs{} ordinary pairs and \evOfficialLocalPairs{} local pairs, and on the three independent external families. The principal audit excludes Protein Data Bank (PDB) matches and protein sequence matches at 40\% identity or greater with at least 70\% coverage of the longer sequence, using the same members for all three models in the audit. To assess how mutation placement affects homomultimer predictions, we held the evaluation members and checkpoints fixed and ran the original single-chain protocol (Frozen) and synchronous substitutions on all homologous chains (All-copy) on the same A10 GPU.

Both controlled models were trained on the absolute-affinity records outside the held-out fold, including records beyond the local mutation cohort. Across five folds, three seeds and two methods, the design gives 30 fits. Both methods used the final checkpoint after 30,000 Adam updates. For each cross-method comparison, we took the intersection of available predictions, including all three seeds for out-of-fold models, then reapplied the local position threshold (Supplementary S1--S2).

\subsection{Metrics and statistical analysis}
For mutation $i$ in landscape $\ell$, the observed effect is $d_{\ell i}=y^{\mathrm{mut}}_{\ell i}-y^{\mathrm{WT}}_{\ell}$, with the predicted effect $\widehat d_{\ell i}$ defined identically. Negative $\Delta$pKd indicates weaker binding. Each position receives equal weight, shared among substitutions. Ordinary $\Delta$MAE is the weighted MAE of $\widehat d-d$. Centered MAE (cMAE) first subtracts each landscape's weighted mean from its observed and predicted effects. We average errors across landscapes within each family, then across families. Absolute MAE averages units within families. Out-of-fold metrics average three separate seed scores. The zero baseline predicts $\Delta$pKd = 0. Position-mean Spearman measures ordering. The position-weighted ratio of predicted to observed standard deviation (SD) measures response amplitude.

Position-mean Spearman is computed within each landscape after averaging predicted and observed effects separately over substitutions at each position. A constant predicted position profile has undefined rank. An SD ratio of one matches experimental spread, and $|\mathrm{SD\ ratio}-1|$ measures amplitude mismatch. The primary local ordinary and centered errors use the same 220 pairs (Supplementary S3).

For endpoint residuals $e_{\mathrm{WT}}$ and $e_{\mathrm{mut}}$, let $c=(e_{\mathrm{mut}}+e_{\mathrm{WT}})/2$ and $q=e_{\mathrm{mut}}-e_{\mathrm{WT}}$. The endpoint squared error per incidence is $(e_{\mathrm{WT}}^2+e_{\mathrm{mut}}^2)/2=c^2+q^2/4$. Within a landscape, mutation squared error partitions into $\mathrm{E}_w[q]^2$ and $\mathrm{E}_w[(q-\mathrm{E}_w[q])^2]$. The centered term is predicted variance plus observed variance minus twice their covariance. Each identity uses common weights and repeated WT incidence in the endpoint decomposition.

Paired intervals resample whole families from fixed predictions. The broad ordinary and endpoint analyses use 2,000 draws. Local analyses share 10,000 draws across metrics and methods. These intervals describe between-family variation for the saved predictions. Intervals are unadjusted. Free energies use $\mathrm{pKd}_{\mathrm{ref}}=-G/(RT\ln 10)$ at 298 K, with $RT\ln 10\approx1.36356$ kcal/mol. mCSM-NA strengthening scores are divided by that denominator. PANTHER uses $-(\mathrm{score}_{\mathrm{mut}}-\mathrm{score}_{\mathrm{WT}})$ in native units.

\subsection{Computational diagnostics, modeling interventions and RAPD}
PRA-Pred endpoints were reconstructed from the executed regression features and original coefficients. Contributions \(\beta_j\Delta x_j\) sum to the unrounded mutation response, distinguishing unchanged inputs, cancellation and display-rounding ties. M2-PRNet diagnostics replayed all eight WT/mutant combinations of atom, residue and image branches through fixed fusion weights. Exact Shapley enumeration~\citep{lundberg2017} used WT states as the baseline and mutant states as replacements. Encoder/readout traces located response attenuation. Mask-only and complete-index alignment changed input organization while retaining physical feature values and checkpoint weights (Supplementary S6–S7).

For each of three seeds, a shared absolute-trained A0 model was cloned into three matched continuations with identical initial parameters, optimizer state, sampling and update budgets. Arm A retained absolute MSE; B added \(0.5(e_{\mathrm{mut}}^2+e_{\mathrm{WT}}^2)\); C added \(0.5(e_{\mathrm{mut}}-e_{\mathrm{WT}})^2\). The auxiliary terms are c\ensuremath{^2} + q\ensuremath{^2}/4 and q\ensuremath{^2}/2: C increases difference weighting while removing the auxiliary shared-error term. This fixed-split experiment is separate from the principal five-fold evaluation. Validation includes 70 pairs, of which 54 local pairs span four held-out families; training-fit diagnostics retain the actual training records and weights (Supplementary S8).

\paragraph{cMAE-selected readouts.}
We tested four linear mutation readouts built from frozen hidden representations, denoised three-dimensional contact changes, their concatenation, and amino-acid composition. Side-chain packing used the released denoising and scoring weights. Each of five outer family folds used inner-fold cMAE to select the regularization setting, with zero response included as a candidate. In 5M3H, the A-chain Arg633$\to$Ala target was repacked in a fixed environment retaining the deposited heavy atoms and $-2e$ formal charge at both background phosphoserines, X:SEP5 and Y:SEP19 (Supplementary S8).

\paragraph{Endpoint-sum potential.}
The SoftKernel single-endpoint interface potential augments each macroscopic affinity prediction with a size-calibrated sum of residue contributions on the fixed-backbone interface $\mathcal{I}_{\mathrm{bb}}(x)$. A protein residue enters this set when a backbone heavy atom lies within 8\,\AA{} of RNA.
\begin{align*}
F(x) &= \widehat y_{\mathrm{macro}}(x)+b_0+b_1|\mathcal{I}_{\mathrm{bb}}(x)|\\
     &\quad+\sum_{r\in\mathcal{I}_{\mathrm{bb}}(x)}
g_\theta\!\left(\phi_r(x)-\bar{\phi}_r(B,R)\right).
\end{align*}
The training-only expectation $\bar{\phi}_r(B,R)$ depends on the fixed backbone and RNA context and includes a fixed apo term. The interface, baseline and size correction are shared across endpoints. Contributions from residues with unchanged features cancel exactly, leaving the local difference. The SoftKernel gate weights residual features by backbone geometry. An endpoint drift penalty constrains the summed correction for each individual training complex (Supplementary S8).

\paragraph{Reference-Anchored Physical Decoupling.}
RAPD separates the absolute output from the mutation predictor. Writing $x_w$ and $x_m$ for the wild type and mutant, respectively,
\begin{align*}
\widehat{y}_{\mathrm{RAPD}}(x) &= \widehat{y}_{\mathrm C}(x),\\
\widehat{\Delta\mathrm{p}K_d}_{\mathrm{RAPD}}(x_w,x_m)
  &= \tfrac{1}{2}\Delta\widehat y_{\mathrm C}
   + \tfrac{1}{2}\Delta\widehat y_{\mathrm A}+\Delta E_{\mathrm{phys}},\\
\Delta E_{\mathrm{phys}}
  &= \beta\,\rapdtheta^{\top}\mathbf D^{-1}
       \bigl[\rapdphi(x_{\mathrm m};x_{\mathrm w})-\rapdphi(x_{\mathrm w};x_{\mathrm w})\bigr].
\end{align*}
Here C and A denote Controlled CoPRA and its additive comparator, and $\mathbf D$ contains feature scales estimated on training data alone. The nine channels capture heavy-atom contacts, $\tanh$-softened clashes, phosphate electrostatics, ribose and base hydrogen-bond proxies, aromatic stacking, net buried surface area, side-chain torsion changes, and site-masked, backbone-conditioned LigandMPNN holo-minus-apo log likelihoods~\citep{dauparas2025ligandmpnn}. For the Npl3 nuclear magnetic resonance (NMR) structure 7QDE, we averaged the feature differences from the first five prespecified models. RNA variant endpoints in the absolute cohort retained their reference protein geometry.

Each outer fold used four inner family folds to select a sign-constrained physical operator. The fit combined centered Huber loss with a pairwise ranking term within landscapes and an $\ell_2$ penalty. On pooled inner predictions, candidates had to satisfy the zero-correction cMAE criterion for the physical branch. Selection then maximized position Spearman, with ties resolved by cMAE, $\beta$ and regularization in that order. Parameters and scale were frozen before outer-fold scoring, and the fixed macro offset was applied after selection. The Decoupled 3D-Ridge and Decoupled AA-only controls share a $0.5\,\Delta\widehat y_{\mathrm C}$ offset and a centered, rank-augmented $\ell_2$-regularized objective. Their matched comparison tests a denoised structural residual representation against amino-acid composition (Supplementary S8).

\textbf{External test.} We reconciled SF3A1, SEPSECS and Npl3 constructs, WT anchors and fixed-backbone inputs with the source studies~\citep{devries2022,puppala2023,moursy2023}. Before prediction we checked provenance, development use and joint sequence-family connectivity against all 771 fitted units (40\% global identity, 70\% coverage). We averaged five fixed fold predictions per endpoint within each seed before differencing, then averaged three separate seed scores. The external test used 24 quantitative pairs and 27 absolute units. Four censored or nonbinding pairs contributed to coverage. Each source panel forms one landscape and one family. Equal-family aggregation gives each panel the same weight (Supplementary S9).

\section{Results}

\subsection{Gains differ across prediction levels}

\begin{figure*}[!t]
\centering\includegraphics[width=153mm]{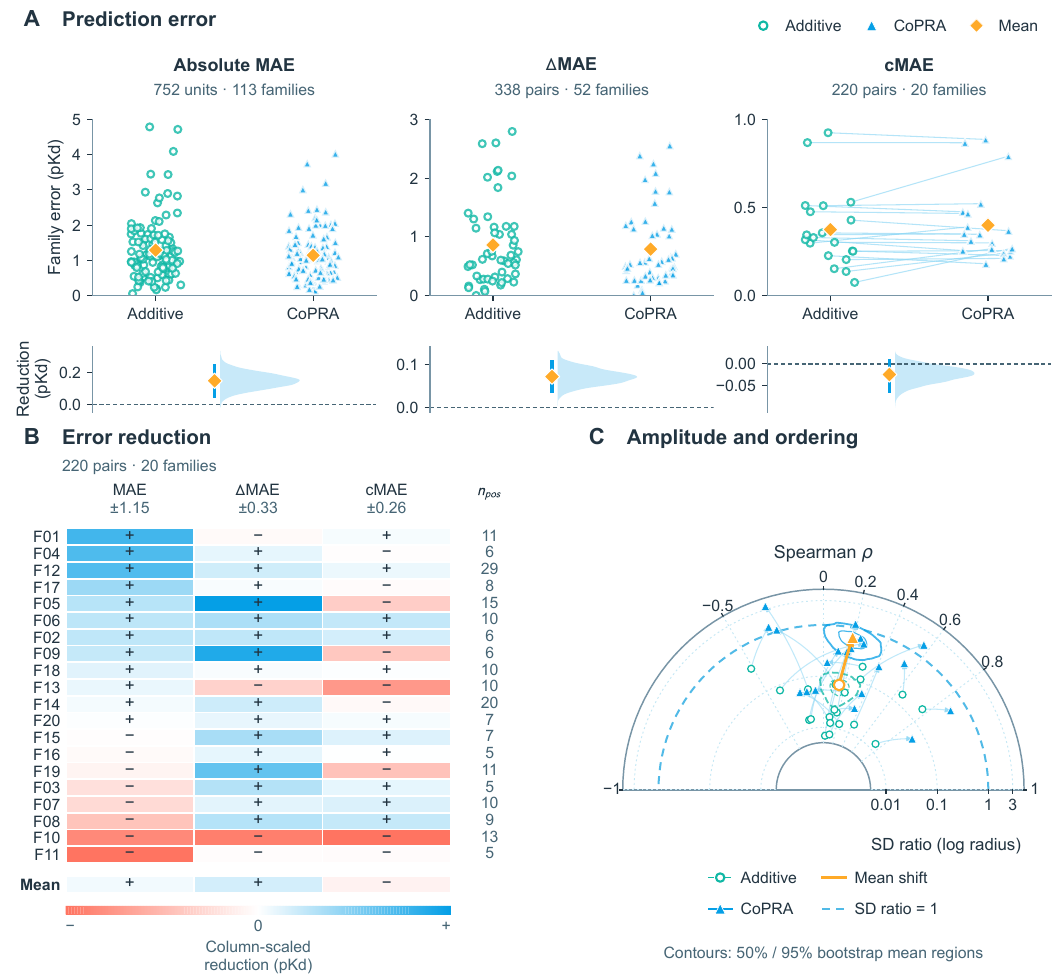}
\caption{\textbf{Controlled CoPRA errors, family reductions and amplitude--ordering.} \textbf{A}, Family errors, equal-family means and 95\% bootstrap intervals for 752 absolute units, 338 ordinary pairs and 220 local pairs. Paired reductions are additive minus Controlled CoPRA. \textbf{B}, Column-scaled reductions in absolute MAE, ordinary $\Delta$MAE and local cMAE across 21 landscapes and 20 families, with position counts and a separate mean row. \textbf{C}, Paired family points and the mean shift in response SD ratio and position Spearman. Contours show the 50\% and 95\% highest-density bootstrap regions. The dashed arc marks an SD ratio of one.}
\label{fig:main1}
\figurealt{Three paired distributions compare Additive with Controlled CoPRA. A family matrix shows heterogeneous reductions, and angle and log radius show ordering and response amplitude.}
\end{figure*}

CoPRA reduced absolute and ordinary mutation errors (Figure~\ref{fig:main1}; Table~\ref{tab:controlled-errors}). On the 220 local pairs, ordinary $\Delta$MAE fell while the paired cMAE contrast spanned zero.

The predicted-to-observed SD ratio showed stronger responses.\ Position-mean Spearman measures ordering, and distance from an SD ratio of one measures amplitude calibration. The paired intervals for ordering and calibration both span zero.

Table~\ref{tab:common-methods} reports gains against matched additive references. In the no-detected-PRA310-overlap subset, Official Released CoPRA had an absolute MAE of \evOfficialAbs{} versus \evMatchedControlAbs{} pKd for matched Controlled CoPRA, an ordinary $\Delta$MAE of \evOfficialOrd{} versus \evMatchedControlOrd{} pKd, and a local cMAE of \evOfficialLocal{} versus \evMatchedControlLocal{} pKd (Table~\ref{tab:rapd}B). The response SD ratio was \evOfficialSD{} versus \evMatchedControlSD{}, and position Spearman was \evOfficialRank{} versus \evMatchedControlRank{}. Severe local compression thus persisted in a released configuration that includes PRI30k complex pretraining. The matched evaluation concerns those published checkpoints and their combined pretraining and fine-tuning history.

\begin{table}[!t]
\begingroup
\centering
\caption{\textbf{Controlled CoPRA errors, local capabilities and paired family uncertainty.}}
\label{tab:controlled-errors}
\fontsize{7.5}{9}\selectfont
\setlength{\tabcolsep}{2.6pt}
\renewcommand{\arraystretch}{1.06}
\begin{tabularx}{\linewidth}{@{}Yrrr>{\raggedleft\arraybackslash}p{21mm}@{}}
\toprule
Measure & CoPRA & Additive & Zero & \shortstack[r]{Paired effect\\{}[95\% interval]} \\
\midrule
\multicolumn{5}{@{}l}{\textbf{A. Broad evaluation: 752 units / 338 ordinary pairs}}\\
Absolute MAE, 752 units & 1.1431 & 1.2898 & --- & --- \\
Linked absolute MAE, 338-pair endpoints & 1.2008 & 1.3345 & --- & --- \\
Ordinary $\Delta$MAE, 338 pairs & 0.7917 & 0.8629 & 0.8816 & \shortstack[r]{0.0713\\{}[0.0362, 0.1068]} \\
\midrule
\multicolumn{5}{@{}l}{\textbf{B. Local evaluation: 220 pairs / 21 landscapes / 20 families}}\\
Linked absolute MAE, 241 units & 1.0593 & 1.1139 & --- & \shortstack[r]{0.0547\\{}[$-$0.1784, 0.2692]} \\
Ordinary $\Delta$MAE & 0.6841 & 0.7426 & 0.7580 & \shortstack[r]{0.0585\\{}[0.0006, 0.1126]} \\
cMAE & 0.3991 & 0.3749 & 0.3787 & \shortstack[r]{$-$0.0243\\{}[$-$0.0635, 0.0088]} \\
Position-mean Spearman & 0.1886 & 0.1481 & --- & \shortstack[r]{0.0405\\{}[$-$0.0822, 0.1645]} \\
Predicted / observed SD & 0.6422 & 0.0709 & 0 & \shortstack[r]{0.5714\\{}[0.2808, 0.9560]} \\
$|\mathrm{SD\ ratio}-1|$ & 0.7707 & 0.9291 & 1 & \shortstack[r]{0.1585\\{}[$-$0.0596, 0.3266]} \\
\bottomrule
\end{tabularx}
\par\smallskip
{\raggedright\fontsize{7.5}{9}\selectfont
Sources: Tables S17, S18 and S21. CoPRA denotes Controlled CoPRA and Zero denotes Zero-change.
Errors use pKd. Effects are additive minus CoPRA for errors and amplitude distance, and CoPRA minus additive for Spearman and SD ratio. Intervals use paired family resampling.
See Supplementary Table S36 for contrasts with zero change.\par}
\endgroup
\end{table}

\begin{table*}[b]
\centering
\caption{\textbf{Error reductions against matched additive references.}}
\label{tab:common-methods}
\begingroup\fontsize{8}{9.5}\selectfont
\setlength{\tabcolsep}{3pt}\renewcommand{\arraystretch}{1.05}
\begin{tabularx}{\textwidth}{@{}Yllrrrrrr@{}}
\toprule
& & & \multicolumn{3}{c}{\textbf{Shared coverage}} & \multicolumn{3}{c}{\textbf{Error reduction (pKd)}} \\
\cmidrule(lr){4-6}\cmidrule(l){7-9}
\textbf{Method} & \textbf{Scope} & \textbf{Ref.} & \textbf{Abs.} & \textbf{Ord.} & \textbf{Loc.} & \textbf{Abs. MAE} & \textbf{Ord. $\Delta$MAE} & \textbf{Loc. cMAE} \\
\midrule
DeePNAP (OOF) & S3 & Seq. add. & 562 & 242 & 161 & $-0.0631$ & $-0.0240$ & $-0.0084$ \\
OmniBioTE official & S4 & CoPRA add. & 736 & 328 & 220 & $+0.6322$ & $-0.0157$ & $-0.0098$ \\
OmniBioTE adapted (OOF) & S4 & CoPRA add. & 736 & 328 & 220 & $-0.1681$ & $-0.0286$ & $-0.0137$ \\
PRdeltaGPred & S6 & CoPRA add. & 286 & 235 & 153 & $-0.4943$ & $-0.0319$ & $-0.0238$ \\
mCSM-NA & S7 & CoPRA add. & --- & 246 & 157 & --- & $-1.2759$ & $-1.3153$ \\
GBNL reconstruction (OOF) & S8 & CoPRA add. & 391 & 149 & 82 & $+0.1131$ & $-0.0163$ & $-0.0091$ \\
SVSBI reconstruction (OOF) & S8 & CoPRA add. & 391 & 149 & 82 & $+0.0623$ & $-0.0202$ & $-0.0048$ \\
ProRB author checkpoint & S8 & CoPRA add. & 391 & 149 & 82 & $-0.0029$ & $-0.0420$ & $-0.0102$ \\
PRA-Pred & S9 & CoPRA add. & 121 & 106 & 78 & $+0.7073$ & $+0.0167$ & $-0.0119$ \\
M2-PRNet, indices aligned & S10 & CoPRA add. & 257 & 211 & 131 & $-0.1065$ & $-0.0117$ & $-0.0018$ \\
Official Released CoPRA & Ov0 & CoPRA add. & \evOfficialEndpoints{} & \evOfficialPairs{} & \evOfficialLocalPairs{} & \evOfficialReductionAbs{} & \evOfficialReductionOrd{} & \evOfficialReductionLocal{} \\
\bottomrule
\end{tabularx}
\par\smallskip
{\raggedright\fontsize{8}{9.5}\selectfont
Each reduction is reference error minus method error on identical members. Positive values favor the listed method. Seq.~add. denotes the sequence additive comparator and CoPRA~add. denotes the CoPRA additive comparator. Coverage counts absolute units, ordinary pairs and local pairs. Scope codes identify the replay subsets. Ov0 denotes the \evOfficialEndpoints{}/\evOfficialPairs{}/\evOfficialLocalPairs{} subset without detected PRA310 overlap. Supplementary S5 and S8 give matched results. Supplementary S5 gives full per-method errors and references, including OmniBioTE's additional sequence-method intersection (Table S3) and PANTHER diagnostics (Table S11).\par}
\endgroup
\end{table*}

Table~\ref{tab:common-methods} reports gains against matched additive references. Official OmniBioTE lowered absolute error on 736 units, with ordinary errors near zero change. PRA-Pred also had low absolute error, with ordinary error close to CoPRA on its scope. M2-PRNet compressed responses (SD ratio 0.0367), as did ProRB, SVSBI and GBNL (0.0515--0.2048). mCSM-NA amplified responses (SD ratio 6.197) and had local cMAE of 1.6998. Supplementary S5 gives complete same-member diagnostics.

\noindent\textbf{PANTHER (native score).} The service gave zero differences for 47/87 pairs and matched the strict direction in 26.4\%. Mean position Spearman was 0.2796 over the four commonly rankable landscapes (31 pairs at 29 positions). Predictions were constant in the fifth eligible landscape. See Supplementary Table S11 for matched reference values.

\subsection{Error reductions concentrate in shared and landscape-mean components}

Endpoint residuals shared by both members cancel when affinities are differenced. On 338 pairs, controlled CoPRA reduced pair-incidence endpoint MSE by 0.4227 pKd\ensuremath{^2}: the shared component contributed 0.3686 and the difference component 0.0541 (Supplementary Figure S19 and Table S17). The shared-component interval crossed zero, whereas the difference-component interval was positive. Thus some improvement reached mutation differences, even though the larger point-estimate contribution concerned shared endpoint placement.

\begin{figure*}[!t]
\centering\includegraphics[width=\textwidth]{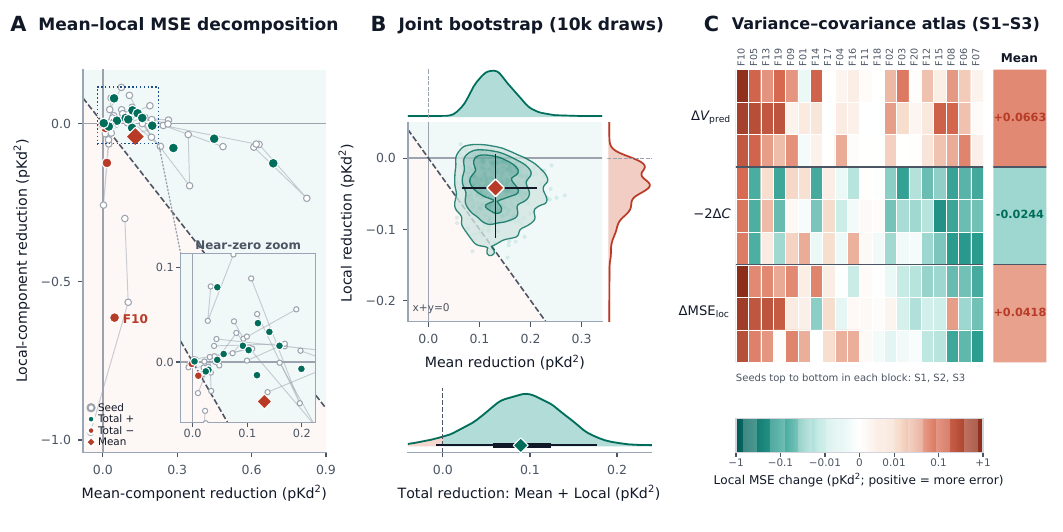}
\caption{\textbf{Landscape-mean and local error decomposition.} \textbf{A}, Family means and linked seed values for additive-minus-CoPRA MSE reductions on 220 pairs in 21 landscapes and 20 families, with a near-zero inset. \textbf{B}, Joint and marginal densities from 10,000 paired family draws, with marginal 95\% intervals and a separate density for total reduction. \textbf{C}, Signed variance, covariance and local-MSE changes by family and seed, with overall means. Units are pKd\textsuperscript{2}. See Supplementary S11 and Figure S17 for the supervision comparison.}
\label{fig:main3}
\figurealt{Linked seeds and family means show mean and local MSE changes, with a near-zero inset. Joint and marginal bootstrap densities show uncertainty. A family-by-seed atlas separates variance, covariance and local MSE.}
\end{figure*}

The second decomposition locates the remaining gain within landscapes. On the same 220 local pairs used for centered accuracy, the landscape-mean term improved by 0.1313 pKd\ensuremath{^2} (95\% interval, 0.0674 to 0.2103), while the centered term worsened by 0.0418 (reduction \ensuremath{-}0.0418; \ensuremath{-}0.1109 to 0.0040). Their sum yielded a total mutation-MSE reduction of 0.0895 (Figure~\ref{fig:main3}A–B). Better average effects accounted for the gain; variation among substitutions had higher mean error. These are matched-member comparisons, with the broader endpoint analysis serving a separate descriptive scope.

Variance–covariance accounting explains why stronger responses did not improve local MSE. Relative to additive, CoPRA increased predicted variance by 0.06626 pKd\ensuremath{^2}, while twice its covariance gain with experiment offset only 0.02443. The remaining 0.04183 is the centered-MSE increase (Figure~\ref{fig:main3}C and Supplementary Table S22).

\subsection{Computation can suppress responses before evaluation}
\begin{figure*}[!t]
\centering\includegraphics[width=\textwidth]{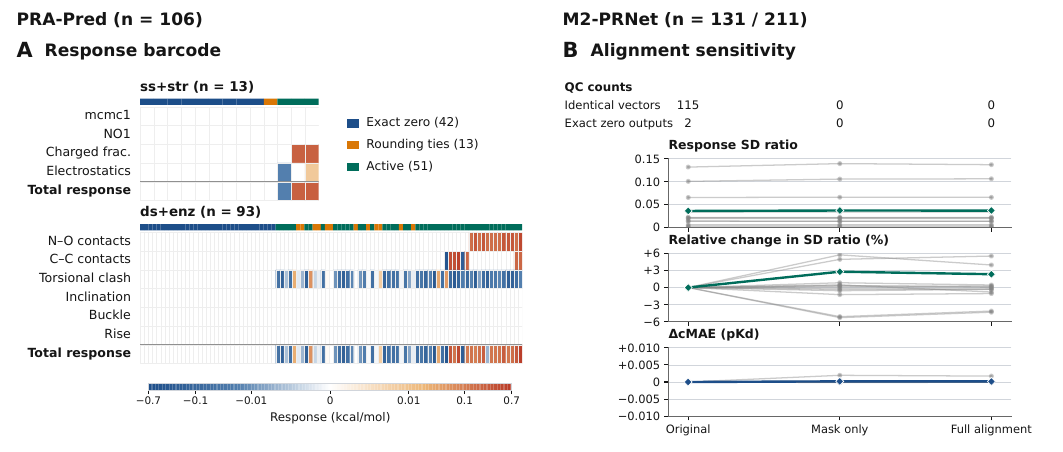}
\caption{\textbf{Mutation responses and alignment sensitivity.}
\textbf{A}, PRA-Pred term contributions and total responses for 106 pairs on a symmetric-log scale. The strips mark exact zeros, rounding-only ties and active responses.
\textbf{B}, M2-PRNet alignment sensitivity. QC counts give identical residue vectors out of 1,055 pair-checkpoint records and exact zero outputs out of 211 ensemble responses. Gray trajectories represent the 12 landscapes and diamonds mark the macro summaries. Relative changes in the SD ratio and differences in cMAE use the original alignment as the baseline.}
\label{fig:main4}
\figurealt{The PRA-Pred barcode distinguishes exact zeros, rounding-only ties and active responses. Alignment tracks show M2-PRNet defect removal, relative change in response amplitude and change in local error across landscapes.}
\end{figure*}

All 42 exact unrounded zeros among 106 PRA-Pred pairs had unchanged effective regression features. Rounding endpoints to two decimals added 13 ties (Figure~\ref{fig:main4}A). In 2C4R F57A, the mutation removes a contact whose energy was already clipped to zero and leaves the charge category unchanged, so every effective input stays the same. Supplementary S6 and Table S24 trace the fixed-backbone RNA geometry terms and the retained NO/NO1 naming behavior.

\noindent\textbf{M2-PRNet input alignment}\par\medskip
In the original reconstructed M2-PRNet path, local SD ratios were 0.0249--0.0680 across checkpoints and fell further with averaging (Supplementary Figure S14A). The atom branch supplied 99.5\% of summed mean absolute Shapley contributions, with little cancellation. Encoder/readout tracing found small residue displacement and weak downstream sensitivity, including weak coupling from image changes (Supplementary S7 and Figure S14).

The original adapter's indexing left 67/211 mutation slots inactive relative to the RNA-first network layout. With pairs and checkpoint weights fixed, both mask-only and complete alignment eliminated identical WT/mutant residue vectors and both exact ensemble zeros (Figure~\ref{fig:main4}B). From the original to complete path, the local SD ratio rose from 0.0358 to 0.0367, while family-macro cMAE was nearly identical across original, mask-only and complete paths (0.3405, 0.3407 and 0.3407 pKd). Atom contributions stayed above 99\%. This comparison applies to the reconstructed input paths documented in Supplementary S2 and S7.

\begin{figure*}[!t]
\centering\includegraphics[width=150mm]{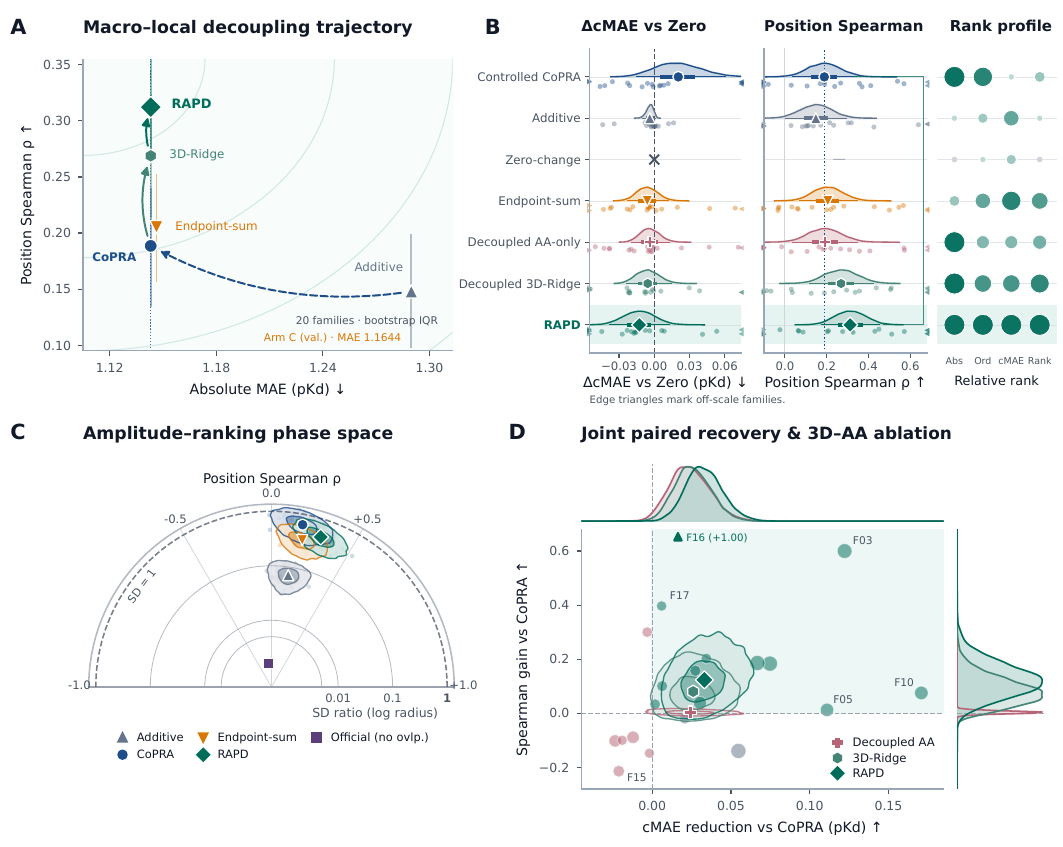}
\caption{\textbf{Reference-anchored recovery on 220 pairs across 20 unseen families.} \textbf{A}, Model trajectory in absolute MAE and position Spearman. \textbf{B}, Rainclouds show paired cMAE differences from Zero-change and position Spearman from 10,000 family draws. Bubble area and saturation encode relative ranks on four metrics. \textbf{C}, Semicircular phase space maps log response amplitude and rank, with 50\% and 95\% bootstrap density regions. The official point uses the subset without detected PRA310 overlap. \textbf{D}, Family-level RAPD recovery and joint bootstrap densities compare decoupled 3D and amino-acid-only readouts with Controlled CoPRA.}\label{fig:rapd}
\figurealt{The model trajectory, rainclouds and rank bubbles compare global error and local recovery. A semicircular density maps response amplitude and ranking. The paired family scatter shows cMAE reduction and Spearman gain relative to Controlled CoPRA.}
\end{figure*}
\subsection{Why difference losses, cMAE-only selection and single-endpoint potentials fail to transfer}
In the three-arm continuation, increasing the weight on endpoint differences improved ordinary validation $\Delta$MAE from 0.7219 to 0.6963 pKd relative to the endpoint-supervised arm. The mean component drove this gain (Supplementary Figures S15 and S17). Training cMAE for the endpoint and difference arms was \evTrainLocalRange{} pKd, with position Spearman \evTrainRankRange{}. In the four held-out families, cMAE rose to \evHoldLocalRange{} pKd and Spearman fell to \evHoldRankRange{} (Supplementary S8). The gap shows that the continuation captured local variation among familiar families, with substantially weaker ordering in the held-out population.

The cMAE-selected hidden, contact, hybrid and composition readouts all chose zero response in each of the five outer folds (SD ratio \evZeroSD{}, local cMAE \evZeroLocal{} pKd). After removing landscape means, the squared-error decomposition shows how higher response variance raises error unless covariance with experiment offsets it. In this family-held-out comparison, the absolute-error criterion likewise favored zero response, consistent with weak transferable local signal.

\begin{table*}[!b]
\caption{\textbf{Modeling interventions and reference-anchored decoupling across matched cohorts.}}
\label{tab:rapd}
\centering
\begingroup
\scriptsize
\setlength{\tabcolsep}{3.25pt}
\begin{tabular*}{\textwidth}{@{\extracolsep{\fill}}>{\raggedright\arraybackslash}p{29mm}rrrrr@{}}\toprule
\textbf{Model} & Abs. MAE & Ord. $\Delta$MAE & Local cMAE & Pos. Spearman & SD ratio\\\midrule
\multicolumn{6}{@{}l}{\textbf{A. Principal full cohort} (752 absolute units, 338 ordinary pairs, 220 local pairs, 20 local families)}\\[1.5pt]
Controlled CoPRA (reference) & $1.1431$ & $0.7917$ & $0.3991$ & $0.1886$ & $0.6422$\\[0.75pt]
Additive comparator & {\fontsize{7.15}{9}\selectfont 1.2898 [+0.0545, +0.2424]} & {\fontsize{7.15}{9}\selectfont 0.8629 [+0.0362, +0.1068]} & {\fontsize{7.15}{9}\selectfont 0.3749 [−0.0635, +0.0088]} & {\fontsize{7.15}{9}\selectfont 0.1481 [−0.1645, +0.0822]} & {\fontsize{7.15}{9}\selectfont 0.0709 [−0.9560, −0.2808]}\\[0.75pt]
Zero-change reference & --- & {\fontsize{7.15}{9}\selectfont 0.8816 [+0.0509, +0.1284]} & {\fontsize{7.15}{9}\selectfont 0.3787 [−0.0609, +0.0151]} & --- & {\fontsize{7.15}{9}\selectfont 0.0000 [−1.0322, −0.3417]}\\[0.75pt]
cMAE-selected readout\textsuperscript{a} & --- & {\fontsize{7.15}{9}\selectfont 0.8816 [+0.0509, +0.1284]} & {\fontsize{7.15}{9}\selectfont 0.3787 [−0.0609, +0.0151]} & --- & {\fontsize{7.15}{9}\selectfont 0.0000 [−1.0322, −0.3417]}\\[0.75pt]
Endpoint-sum potential & {\fontsize{7.15}{9}\selectfont 1.1464 [−0.0500, +0.0587]} & {\fontsize{7.15}{9}\selectfont 0.8107 [−0.0021, +0.0385]} & {\fontsize{7.15}{9}\selectfont 0.3727 [−0.0516, −0.0050]} & {\fontsize{7.15}{9}\selectfont 0.2052 [−0.0227, +0.0586]} & {\fontsize{7.15}{9}\selectfont 0.3389 [−0.4984, −0.1537]}\\[0.75pt]
Decoupled AA-only & $1.1431$\textsuperscript{\dag} & {\fontsize{7.15}{9}\selectfont 0.8192 [+0.0052, +0.0483]} & {\fontsize{7.15}{9}\selectfont 0.3749 [−0.0499, −0.0018]} & {\fontsize{7.15}{9}\selectfont 0.1919 [−0.0070, +0.0142]} & {\fontsize{7.15}{9}\selectfont 0.3214 [−0.5156, −0.1709]}\\[0.75pt]
Decoupled 3D-Ridge & $1.1431$\textsuperscript{\dag} & {\fontsize{7.15}{9}\selectfont 0.8071 [−0.0050, +0.0348]} & {\fontsize{7.15}{9}\selectfont 0.3731 [−0.0489, −0.0052]} & {\fontsize{7.15}{9}\selectfont 0.2689 [−0.0046, +0.1981]} & {\fontsize{7.15}{9}\selectfont 0.3516 [−0.4693, −0.1488]}\\[0.75pt]
\textbf{RAPD} & $1.1431$\textsuperscript{\dag} & {\fontsize{7.15}{9}\selectfont 0.7704 [−0.0418, +0.0005]} & {\fontsize{7.15}{9}\selectfont 0.3660 [−0.0569, −0.0120]} & {\fontsize{7.15}{9}\selectfont 0.3123 [+0.0109, +0.2596]} & {\fontsize{7.15}{9}\selectfont 0.4700 [−0.3052, −0.0534]}\\[0.75pt]
\midrule
\multicolumn{6}{@{}l}{\textbf{B. Principal subset without detected PRA310 overlap} (95 absolute units, 53 ordinary pairs, 35 local pairs)}\\[1.5pt]
Controlled CoPRA (matched subset) & $1.5154$ & $0.5741$ & $0.4176$ & $0.1553$ & $0.3487$\\[0.75pt]
Official Released CoPRA & {\fontsize{7.15}{9}\selectfont 1.6776 [−0.3334, +0.6904]} & {\fontsize{7.15}{9}\selectfont 0.6304 [+0.0144, +0.1010]} & {\fontsize{7.15}{9}\selectfont 0.4244 [−0.0112, +0.0358]} & {\fontsize{7.15}{9}\selectfont −0.1065 [−0.8029, +0.3865]} & {\fontsize{7.15}{9}\selectfont 0.0016 [−0.3936, −0.3208]}\\[0.75pt]
\bottomrule\end{tabular*}\par\smallskip
\raggedright Brackets in panels A and B give paired 95\% family-bootstrap intervals for model minus matched Controlled CoPRA. Negative values indicate lower error for MAE metrics, and positive values indicate higher rank correlation or amplitude. Dashes denote undefined or unavailable metrics. Errors are in pKd. \textsuperscript{a}All four readouts (hidden representation, contact, hybrid and amino-acid composition) selected zero response in each outer fold. \textsuperscript{\dag}Identical to the Controlled CoPRA absolute-affinity output by construction. (Continued on next page.)\par\endgroup\end{table*}

\begin{table*}[t]
\ContinuedFloat
\captionsetup{labelformat=continuedtable}
\caption{\textbf{Modeling interventions and reference-anchored decoupling across matched cohorts.}}
\centering
\begingroup
\scriptsize
\setlength{\tabcolsep}{3.5pt}
\begin{tabular*}{\textwidth}{@{\extracolsep{\fill}}p{41mm}rrrrr@{}}\toprule
\textbf{Model} & Abs. MAE & Ord. $\Delta$MAE & Local cMAE & Pos. Spearman & SD ratio\\\midrule
\multicolumn{6}{@{}l}{\textbf{C. Independent external test} (27 absolute units, 24 pairs, 3 source-matched families)}\\[1.5pt]
Controlled CoPRA & \shortstack[r]{$0.5282$\\[-1pt]{\scriptsize $[0.3770,\,0.8289]$}} & \shortstack[r]{$0.6995$\\[-1pt]{\scriptsize $[0.3982,\,1.0317]$}} & \shortstack[r]{$0.3761$\\[-1pt]{\scriptsize $[0.3357,\,0.3978]$}} & \shortstack[r]{$0.1454$\\[-1pt]{\scriptsize $[-0.2619,\,0.4095]$}} & \shortstack[r]{$0.3900$\\[-1pt]{\scriptsize $[0.3067,\,0.4692]$}}\\[0.75pt]
Official Released CoPRA & \shortstack[r]{$1.4606$\\[-1pt]{\scriptsize $[0.6353,\,1.8769]$}} & \shortstack[r]{$0.9320$\\[-1pt]{\scriptsize $[0.4945,\,1.3977]$}} & \shortstack[r]{$0.4241$\\[-1pt]{\scriptsize $[0.3704,\,0.4902]$}} & \shortstack[r]{$-0.2112$\\[-1pt]{\scriptsize $[-0.4857,\,0.1071]$}} & \shortstack[r]{$0.0149$\\[-1pt]{\scriptsize $[0.0070,\,0.0222]$}}\\[0.75pt]
Additive comparator & \shortstack[r]{$0.6932$\\[-1pt]{\scriptsize $[0.4483,\,0.9397]$}} & \shortstack[r]{$0.8845$\\[-1pt]{\scriptsize $[0.4866,\,1.2998]$}} & \shortstack[r]{$0.4198$\\[-1pt]{\scriptsize $[0.3553,\,0.4869]$}} & \shortstack[r]{$0.2208$\\[-1pt]{\scriptsize $[-0.2571,\,0.7024]$}} & \shortstack[r]{$0.0467$\\[-1pt]{\scriptsize $[0.0173,\,0.0786]$}}\\[0.75pt]
\shortstack[l]{Zero-change /\\cMAE-selected readout} & --- & \shortstack[r]{$0.9223$\\[-1pt]{\scriptsize $[0.4926,\,1.3803]$}} & \shortstack[r]{$0.4230$\\[-1pt]{\scriptsize $[0.3688,\,0.4892]$}} & --- & \shortstack[r]{$0.0000$\\[-1pt]{\scriptsize $[0.0000,\,0.0000]$}}\\[0.75pt]
Endpoint-sum potential & \shortstack[r]{$0.5849$\\[-1pt]{\scriptsize $[0.4029,\,0.8701]$}} & \shortstack[r]{$0.7860$\\[-1pt]{\scriptsize $[0.4369,\,1.1652]$}} & \shortstack[r]{$0.3935$\\[-1pt]{\scriptsize $[0.3642,\,0.4383]$}} & \shortstack[r]{$0.1811$\\[-1pt]{\scriptsize $[-0.1190,\,0.3905]$}} & \shortstack[r]{$0.2023$\\[-1pt]{\scriptsize $[0.1613,\,0.2429]$}}\\[0.75pt]
Decoupled AA-only & \shortstack[r]{$0.5282$\\[-1pt]{\scriptsize $[0.3770,\,0.8289]$}} & \shortstack[r]{$0.7994$\\[-1pt]{\scriptsize $[0.4347,\,1.2005]$}} & \shortstack[r]{$0.3959$\\[-1pt]{\scriptsize $[0.3732,\,0.4399]$}} & \shortstack[r]{$0.1403$\\[-1pt]{\scriptsize $[-0.2619,\,0.4095]$}} & \shortstack[r]{$0.1951$\\[-1pt]{\scriptsize $[0.1535,\,0.2344]$}}\\[0.75pt]
Decoupled 3D-Ridge & \shortstack[r]{$0.5282$\\[-1pt]{\scriptsize $[0.3770,\,0.8289]$}} & \shortstack[r]{$0.7895$\\[-1pt]{\scriptsize $[0.4270,\,1.2200]$}} & \shortstack[r]{$0.4073$\\[-1pt]{\scriptsize $[0.3775,\,0.4301]$}} & \shortstack[r]{$0.1560$\\[-1pt]{\scriptsize $[-0.4881,\,0.5810]$}} & \shortstack[r]{$0.2358$\\[-1pt]{\scriptsize $[0.1692,\,0.3203]$}}\\[0.75pt]
\textbf{RAPD} & \shortstack[r]{$0.5282$\\[-1pt]{\scriptsize $[0.3770,\,0.8289]$}} & \shortstack[r]{$0.7207$\\[-1pt]{\scriptsize $[0.4111,\,1.1416]$}} & \shortstack[r]{$0.4151$\\[-1pt]{\scriptsize $[0.4075,\,0.4241]$}} & \shortstack[r]{$0.0950$\\[-1pt]{\scriptsize $[-0.3929,\,0.3827]$}} & \shortstack[r]{$0.2677$\\[-1pt]{\scriptsize $[0.2231,\,0.3071]$}}\\[0.75pt]
\bottomrule\end{tabular*}\par\smallskip
\raggedright (Continued) Panel C gives equally weighted family means with observed minima and maxima in brackets. Dashes denote undefined or unavailable metrics. Errors are in pKd.\par\endgroup\end{table*}

{\sloppy The Endpoint-sum potential satisfied the endpoint-difference identity to within \evIdentityBound{} pKd. The difficulty arises at the single-complex scale, where the correction sums all interface terms (median \evInterfaceMedian{} residues across the frozen principal endpoints). Site weights large enough to express a single substitution can shift that sum across complexes, so the zero-drift constraint attenuates the local response. Much of the cMAE improvement came from the equal-weight macroscopic mixture (\evMacroLocal{} pKd), with the SoftKernel potential reaching \evUoneLocal{} pKd and rank \evUoneRank{} (Figure~\ref{fig:rapd}A). On the official affinity backbone, its external mean SD ratio and Spearman were \evUoneOfficialExtSD{} and \evUoneOfficialExtRank{}, respectively (Supplementary S8).\par}

\subsection{RAPD recovers unseen-family local accuracy and ranking}
RAPD retained the controlled absolute predictions exactly (MAE \evRapdAbs{} pKd). Its ordinary $\Delta$MAE across \evPrimaryPairs{} pairs was \evRapdOrd{} pKd, corresponding to a paired change of \evRapdDiffOrd{} pKd. Across \evPrimaryLocalPairs{} local pairs, RAPD recorded the lowest cMAE among the full-cohort models in Table~\ref{tab:rapd}A at \evRapdLocal{} pKd, below both the additive comparator (\evAdditiveLocal{}) and the zero-change reference (\evZeroLocal{}). The paired cMAE change from Controlled CoPRA was \evRapdDiffLocal{} pKd. Position Spearman increased from \evControlRank{} to \evRapdRank{} (paired change \evRapdDiffRank{}), while the SD ratio was \evRapdSD{} (Figure~\ref{fig:rapd}B--D).

Under the common half-CoPRA offset and regularized objective, replacing amino-acid composition with denoised 3D features reduced ordinary $\Delta$MAE from \evAAOrd{} to \evGeometryOrd{} pKd (paired change \evGeometryDiffOrd{} pKd). Local cMAE moved from \evAALocal{} to \evGeometryLocal{} pKd, and position Spearman from \evAARank{} to \evGeometryRank{} (paired change \evGeometryDiffRank{}). The ordinary-error interval supports a contribution from structural information beyond residue identity. The ranking interval spans zero, and RAPD additionally changes the macro mixture and selection rule.

\subsection{Family heterogeneity and synchronous homomultimer substitutions}
CoPRA's local errors varied across constructs and families. Hfq, for example, combines relatively strong ordering with compressed response spread (Supplementary S1). In the matched A10 sensitivity analysis, replacing all homologous copies reduced ordinary $\Delta$MAE on the local set from \evFrozenOrd{} to \evSyncedOrd{} pKd and increased the response SD ratio from \evFrozenSD{} to \evSyncedSD{} (paired interval \evSyncedSDInterval{}). Local cMAE moved from \evFrozenLocal{} to \evSyncedLocal{} pKd, and the position-Spearman change was \evSyncedRankDiff{}. The result links copy assignment to predicted amplitude while leaving its ordering effect uncertain. Removing the complete 1JBR landscape, where mutated-copy correspondence remains unresolved, retained the cMAE pattern on 209 pairs (0.4018 pKd for Controlled CoPRA and 0.3758 pKd for the additive comparator; Supplementary S10).

\subsection{Independent external transfer and structural boundaries}
\begin{figure*}[t]
\centering\includegraphics[width=\textwidth]{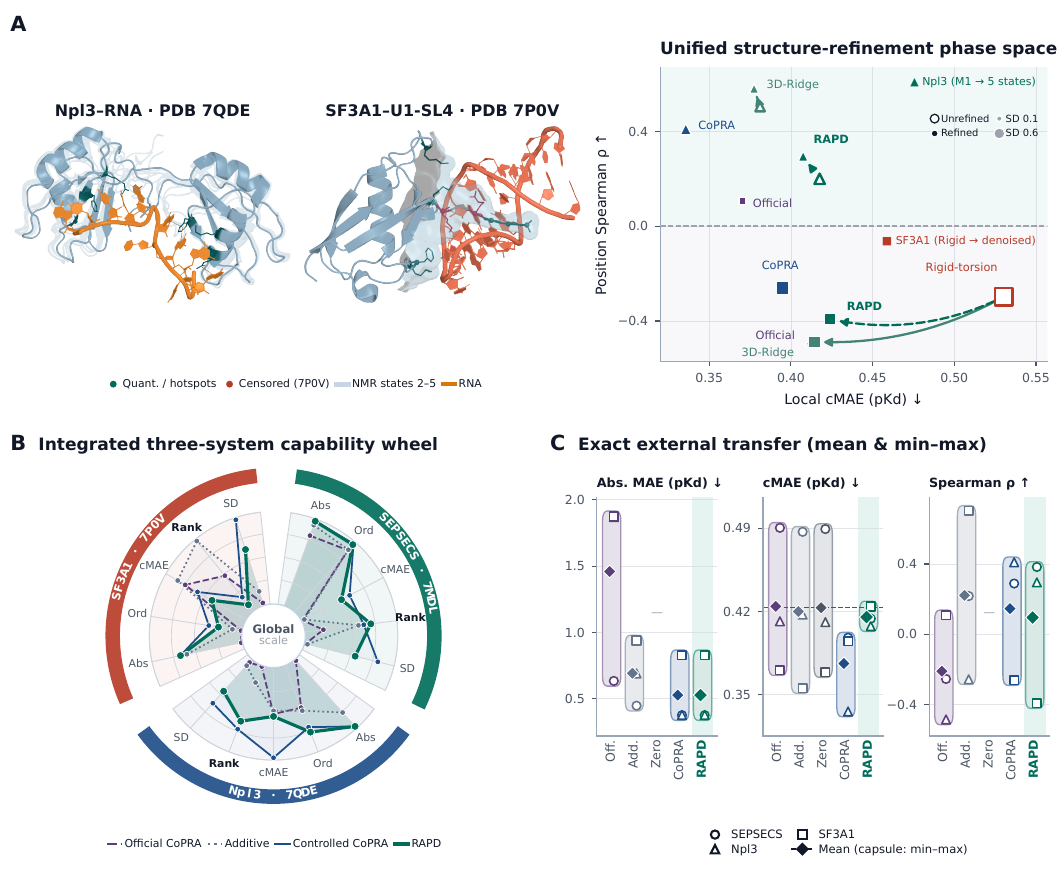}
\caption{\textbf{Structure refinement and independent external transfer.} \textbf{A}, Five superposed WT Npl3 conformers (7QDE), the SF3A1--U1-SL4 interface (7P0V) with quantitative and censored sites, and refinement trajectories in local cMAE and position Spearman. \textbf{B}, Three sectors show five normalized performance and response metrics on a shared scale. \textbf{C}, Open symbols show individual family values, diamonds the equal-family means and rounded capsules the observed min--max ranges. Supplementary Figure E1 extends the capsule display to all five metrics.}\label{fig:transfer}
\figurealt{The Npl3 ensemble and SF3A1 interface sit beside a phase plot of refinement. A five-metric capability wheel summarizes the three external systems. Rounded capsules show observed ranges with family points and equal-family means.}
\end{figure*}

Across \evExternalEndpoints{} absolute units and \evExternalPairs{} quantitative pairs, Controlled CoPRA reduced the equal-family mean absolute MAE from \evAdditiveExtAbs{} to \evControlExtAbs{} pKd and ordinary $\Delta$MAE from \evAdditiveExtOrd{} to \evControlExtOrd{} pKd relative to the additive comparator. Local cMAE fell from \evAdditiveExtLocal{} to \evControlExtLocal{} pKd. Family changes varied: additive-minus-controlled local error was \evControlReductionSep{} pKd for SEPSECS, \evControlReductionNpl{} for Npl3, and \evControlReductionSf{} for SF3A1, where position Spearman was \evControlSfRank{} versus \evAdditiveSfRank{}. Figure~\ref{fig:transfer}B, C summarizes the three families. Individual estimates, observed ranges and three-family means also appear in Table~\ref{tab:rapd}C and Supplementary S9.

The official ensemble returned an equal-family absolute MAE of \evOfficialExtAbs{} pKd and a local cMAE of \evOfficialExtLocal{} pKd, with a response SD ratio of \evOfficialExtSD{} and position Spearman of \evOfficialExtRank{}. RAPD retained the controlled absolute MAE (\evRapdExtAbs{} pKd), with ordinary $\Delta$MAE \evRapdExtOrd{} pKd, local cMAE \evRapdExtLocal{} pKd and position Spearman \evRapdExtRank{}. Its local error was below the zero-change (\evZeroExtLocal{}), additive (\evAdditiveExtLocal{}) and official (\evOfficialExtLocal{}) references, whereas Controlled CoPRA retained a lower mean local error (\evControlExtLocal{} pKd) and higher mean rank correlation (\evControlExtRank{}) than RAPD (Figure~\ref{fig:transfer}C).

Structural comparisons indicated different sensitivities across interfaces. Averaging five 7QDE conformers for Npl3 reduced Decoupled 3D-Ridge cMAE from \evCloudNplSingleLocal{} to \evCloudNplMeanLocal{} pKd and increased its Spearman from \evCloudNplSingleRank{} to \evCloudNplMeanRank{}. The corresponding RAPD cMAE fell from \evRapdNplSingleLocal{} to \evRapdNplMeanLocal{} pKd. For SF3A1--U1-SL4 (7P0V), the rigid-torsion placement and denoised side-chain packing configurations returned cMAEs of \evOldSfLocal{} and \evCloudSfLocal{} pKd, respectively. Repacking changes the steric environment that enters the structural readout, and the archived comparison also changes the feature and fitting protocol. The solvent-exposed UBL loop, which includes sites for three censored substitutions, is a specific setting in which flexible structural sampling may matter for transfer from a static backbone (Figure~\ref{fig:transfer}A).

\section{Discussion}
The two decompositions explain why a predictor can improve aggregate affinity and mutation-effect error without improving the ability to rank substitutions. Shared residuals can be large in the endpoint task and cancel exactly when matched endpoints are differenced. Within a mutation landscape, improving the mean effect can lower ordinary mutation error while leaving the centered response inaccurate. The local squared-error accounting identifies the further requirement that the gain in covariance with experiment offset increased prediction variance. The eleven-family method audit and the overlap-screened analysis of Official Released CoPRA place these distinctions in the context of current affinity predictors.

The interventions locate the bottleneck at three different stages. Increasing the weight on differences made the training-family response easier to recover, yet its ordering transferred poorly. Selecting exclusively by cMAE rewarded weak response when the readout had little transferable signal. A single-endpoint potential had a further scale problem, because the same terms encode a local contrast after differencing and a sum over the whole interface for absolute prediction. RAPD retains the multimodal affinity scale while estimating a low-dimensional physical contrast on paired inputs anchored to the same wild type. The held-out-family gains support this separation, and the external results identify interfaces for which its benefit remains context dependent.

For evaluating mutation tools, cMAE, position ordering and the SD ratio together distinguish local accuracy from attenuated or misordered responses. Each captures a failure mode that can be obscured by absolute error or by the ordinary mutation aggregate. Construct-to-input correspondence determines which comparison is interpretable, particularly when the measured oligomer or RNA differs from the modeled assembly. The Npl3 conformer sensitivity further shows why the structural representation should be considered alongside model choice when evaluating a predictor for a particular construct.

\paragraph{Limitations.}
First, the released CoPRA weights couple PRI30k complex pretraining with PRA310 fine-tuning, so isolating either stage would require intermediate checkpoints. Second, third-party comparisons use method-specific common members, and the same-device homomultimer test changes sequence placement on a fixed backbone. The 1JBR exclusion addresses one unresolved copy assignment. Third, static protein backbones and implicit solvent omit flexible loop rearrangements and explicit water and ion redistribution, features that can influence interfaces such as SF3A1--U1-SL4. Fourth, the \evLocalFamilies{} principal local families and three external families define the empirical scope of the comparison. Flexible-backbone relaxation and high-throughput mutation pretraining offer ways to test generalization across a wider range of binding systems.

\declarationheading{Data availability}
The primary binding-affinity measurements were obtained from \href{https://web.iitm.ac.in/bioinfo2/pronab/}{ProNAB} and are subject to the provider's terms. Supplementary Section S9 identifies the measurement sources for the independent external test. \ReleaseArchiveText{} Access to third-party model checkpoints is governed by the respective providers' terms.

\declareifset{Author contributions}{\AuthorContributionsText}
\declareifset{Funding}{\FundingText}
\declareifset{Acknowledgements}{\AcknowledgementsText}
\declareifset{Conflict of interest}{\ConflictOfInterestText}
\begin{samepage}
\declareifset{AI assistance}{\AIAssistanceText}
\end{samepage}

\section*{Key Points}
\begin{itemize}\setlength{\itemsep}{2pt}
\item Comparison of eleven method families and a separate audit of Official Released CoPRA traced most gains to shared endpoint placement and landscape means.
\item PRA-Pred feature tracing and fixed-weight M2-PRNet controls located response compression within their prediction pipelines.
\item Difference-weighted supervision transferred poorly, selection by cMAE favored zero responses, and interface-wide endpoint sums diluted site-level effects.
\item RAPD reached a cMAE of \evRapdLocal{} pKd and position Spearman of \evRapdRank{} in \evLocalFamilies{} unseen families. Transfer across three independent families varied with the structural input and binding interface.
\end{itemize}

\section*{Biographical Note}
Shangkun Sima is an undergraduate student in the School of Mathematics and Statistics at Lanzhou University, China. His research focuses on statistical modeling, machine learning and quantitative evaluation of biomolecular interactions.

Tongfei Shen is a PhD candidate affiliated with the School of Information Science and Engineering, Qingdao Huanghai University, and the Department of Computer Science, City University of Hong Kong. His research interests include computational biology and biomolecular data curation.

Tianchi Lu is a postdoctoral fellow at The Hong Kong Polytechnic University, Hong Kong SAR, China. His research focuses on machine learning and structural bioinformatics for protein--nucleic acid recognition.

\end{document}